\documentclass[10pt,twocolumn,letterpaper]{article}

\usepackage{iccv}
\usepackage{times}
\usepackage{epsfig}
\usepackage{graphicx}
\usepackage{amsmath}
\usepackage{amssymb}
\usepackage{multirow}
\usepackage{makecell}
\usepackage{threeparttable}
\usepackage{booktabs}
\usepackage{stfloats}
\usepackage{url}
\usepackage{footnote}
\usepackage{comment}
\usepackage{subfig}

\usepackage{graphicx}
\usepackage{amsmath,amssymb} 
\usepackage{tabularx}

\usepackage[british,UKenglish,USenglish,english,american]{babel}

\usepackage[toc,page]{appendix}

\usepackage[bookmarks=false,colorlinks=true,linkcolor=black,citecolor=black,filecolor=black,urlcolor=black]{hyperref}

\newcommand{\tn}[1]{\footnotesize{#1}}

\newcolumntype{x}{>\small c}

\iccvfinalcopy 

\def\iccvPaperID{***} 

\begin{document}

\title{BoxSup: Exploiting Bounding Boxes to Supervise Convolutional Networks \\ for Semantic Segmentation}

\author{Jifeng Dai \qquad\qquad Kaiming He \qquad\qquad Jian Sun \vspace{8pt}\\
Microsoft Research\\
{\tt\small \{jifdai,kahe,jiansun\}@microsoft.com}
}

\maketitle

\begin{abstract}
Recent leading approaches to semantic segmentation rely on deep convolutional networks trained with
human-annotated, pixel-level segmentation masks. Such pixel-accurate supervision demands expensive labeling effort and limits the performance of deep networks that usually benefit from more training data. In this paper, we propose a method that achieves competitive accuracy but only requires easily obtained bounding box annotations. The basic idea is to iterate between automatically generating region proposals and training convolutional networks. These two steps gradually recover segmentation masks for improving the networks, and vise versa.
Our method, called ``BoxSup'', produces competitive results (\eg, 62.0\% mAP for validation) supervised by boxes only, on par with strong baselines (\eg, 63.8\% mAP) fully supervised by masks under the same setting.
By leveraging a large amount of bounding boxes, BoxSup further unleashes the power of deep convolutional networks and yields state-of-the-art results on PASCAL VOC 2012 and PASCAL-CONTEXT \cite{mottaghi2014role}.
\end{abstract}

\section{Introduction}

In the past few months, tremendous progress has been made in the field of semantic segmentation \cite{hariharan2014simultaneous,Long2015,Hariharan2015,Dai2015,Chen2015,mostajabi2014feedforward}. Deep convolutional neural networks (CNNs) \cite{lecun1989backpropagation,krizhevsky2012imagenet} that play as rich hierarchical feature extractors are a key to these methods. These networks are trained on large-scale datasets \cite{deng2009imagenet,Russakovsky2014} as classifiers, and transferred to the semantic segmentation tasks based on the annotated segmentation masks as supervision.

\begin{figure*}[t]
	\centering
	\includegraphics[width=.9\linewidth]{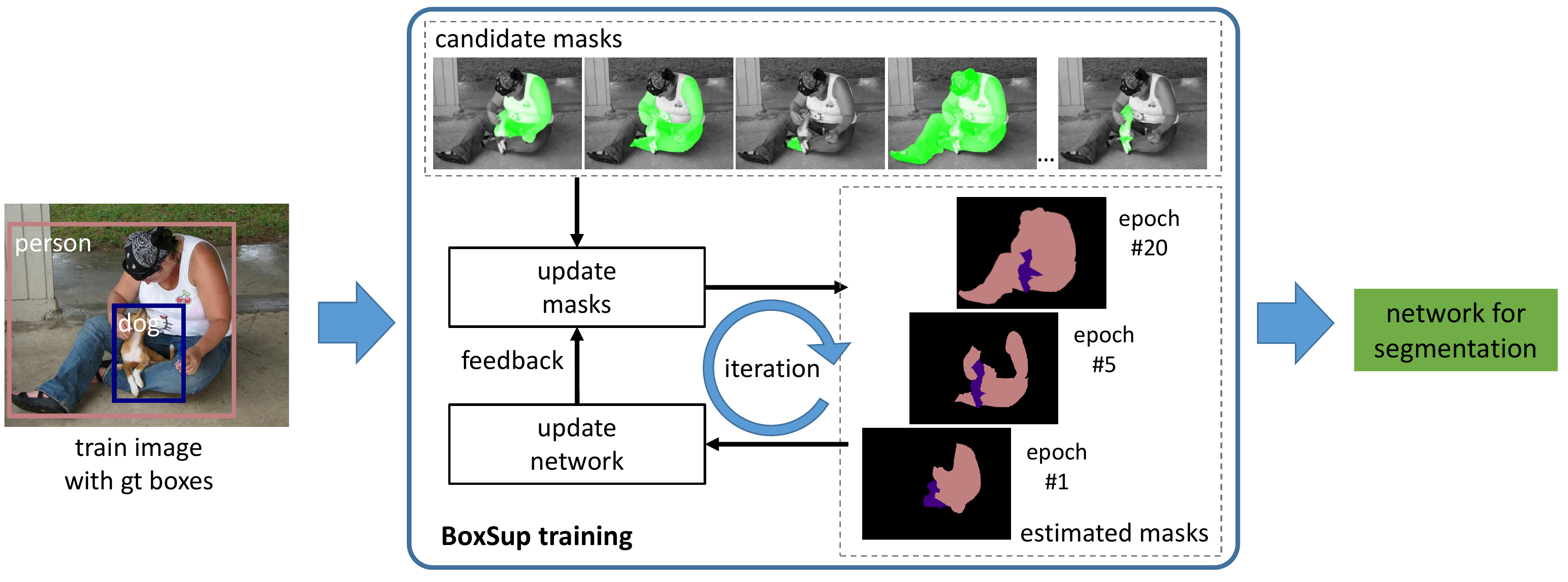}
	\caption{Overview of our training approach supervised by bounding boxes.}
	\label{fig:outline}
\end{figure*}

But pixel-level mask annotations are time-consuming, frustrating, and in the end commercially expensive to obtain.
According to the annotation report of the large-scale Microsoft COCO dataset \cite{Lin2014}, the workload of labeling segmentation masks is more than 15 times heavier than that of spotting object locations. Further, the crowdsourcing annotators need to be specially trained for the tedious and difficult task of labeling per-pixel masks. These facts limit the amount of available segmentation mask annotations, and thus hinder the performance of CNNs that in general desire large-scale data for training. On the contrary, bounding box annotations are more economical than masks. There have already existed a large number of available box-level annotations in datasets like PASCAL VOC 2007\footnote{The PASCAL VOC 2007 dataset only has bounding box annotations.} \cite{everingham2010pascal} and ImageNet \cite{Russakovsky2014}. Though these box-level annotations are less precise than pixel-level masks, their amount may help improve training deep networks for semantic segmentation.

In addition, current leading approaches have not fully utilized the detailed pixel-level annotations. For example, in the Convolutional Feature Masking (CFM) method \cite{Dai2015}, the fine-resolution masks are used to generate very low-resolution (\eg, $6 \times 6$) masks on the feature maps. In the Fully Convolutional Network (FCN) method \cite{Long2015}, the network predictions are regressed to the ground-truth masks using a large stride (\eg, 8 pixels). These methods yield competitive results without explicitly harnessing the finer masks. If we consider the box-level annotations as very coarse masks, can we still retain comparably good results without using the segmentation masks?

In this work, we investigate bounding box annotations as an alternative or extra source of supervision to train convolutional networks for semantic segmentation\footnote{The idea of using bounding box annotations for CNN-based semantic segmentation is developed concurrently and independently in \cite{papandreou2015weakly}. We also compare with the results of \cite{papandreou2015weakly}.}.
We resort to unsupervised region proposal methods \cite{uijlings2013selective,arbelaez2014multiscale} to generate candidate segmentation masks. The convolutional network is trained under the supervision of these approximate masks. The updated network in turn improves the estimated masks used for training. This process is iterated. Although the masks are coarse at the beginning, they are gradually improved and then provide useful information for network training. Fig.~\ref{fig:outline} illustrates our training algorithm.

We extensively evaluate our method, called ``BoxSup'', on the PASCAL segmentation benchmarks \cite{everingham2010pascal,mottaghi2014role}. Our box-supervised (\ie, using bounding box annotations) method shows a graceful degradation compared with its mask-supervised (\ie, using mask annotations) counterpart. As such, our method waives the requirement of pixel-level masks for training.
Further, our semi-supervised variant in which 9/10 mask annotations are replaced with bounding box annotations yields comparable accuracy with the fully mask-supervised counterpart.
This suggests that we may save expensive labeling effort by using bounding box annotations dominantly.
Moreover, our method makes it possible to harness the large number of available box annotations to improve the mask-supervised results.
Using the limited provided mask annotations and extra large-scale bounding box annotations,
our method achieves state-of-the-art results on both PASCAL VOC 2012 and PASCAL-CONTEXT \cite{mottaghi2014role} benchmarks.

Why can a large amount of bounding boxes help improve convolutional networks? Our error analysis reveals that a BoxSup model trained with a large set of boxes effectively increases the object \emph{recognition} accuracy (the accuracy in the middle of an object), and its improvement on object boundaries is secondary. Though a box is too coarse to contain detailed segmentation information, it provides an instance for learning to distinguish object categories.
The large-scale object instances improve the feature quality of the learned convolutional networks, and thus impact the overall performance for semantic segmentation.

\section{Related Work}

Deep convolutional networks in general have better accuracy with the growing size of training data, as is evidenced in \cite{krizhevsky2012imagenet,Zeiler2014}.
The ImageNet classification dataset \cite{Russakovsky2014} is one of the largest datasets with quality labels, but the current available datasets for object detection, semantic segmentation, and many other vision tasks mostly have orders of magnitudes fewer labeled samples. The milestone work of R-CNN \cite{Girshick2014} proposes to pre-train deep networks as classifiers on the large-scale ImageNet dataset and go on training (fine-tuning) them for other tasks that have limited number of training data. This transfer learning strategy is widely adopted for object detection \cite{Girshick2014,He2014,Szegedy2015}, semantic segmentation \cite{hariharan2014simultaneous,Long2015,Hariharan2015,Dai2015,Chen2015,mostajabi2014feedforward},
visual tracking \cite{Wang2015}, and other visual recognition tasks. With the continuously improving deep convolutional models \cite{Zeiler2014,Sermanet2014,Chatfield2014,He2014,Simonyan2015,Szegedy2015,He2015}, the accuracy of these vision tasks also improves thanks to the more powerful generic features learned from large-scale datasets.

Although pre-training partially relieves the problem of limited data, the amount of the task-specific data for fine-tuning still matters. In \cite{agrawal2014analyzing}, it has been found that augmenting the object detection training set by combining the VOC 2007 and VOC 2012 sets improves object detection accuracy compared with using VOC 2007 only. In \cite{Liang2014}, the training set for object detection is augmented by visual tracking results obtained from videos and improves detection accuracy. 
These experiments demonstrate the importance of dataset sizes for task-specific network training.

For semantic segmentation, there have been existing papers \cite{xia2013semantic,guillaumin2014imagenet} that investigate exploiting bounding box annotations instead of masks. But the box-level annotations have not been used to supervised deep convolutional networks in those works.

\begin{figure*}[t]
	\centering
	\includegraphics[width=1.0\linewidth]{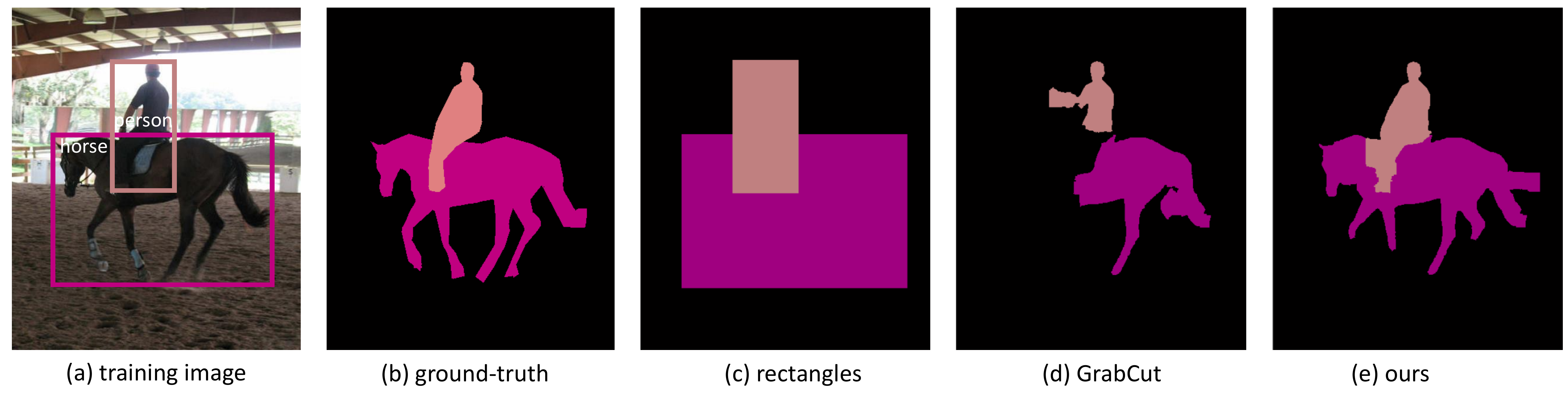}
	\caption{Segmentation masks used as supervision. (a) A training image. (b) Ground-truth. (c) Each box is na\"{\i}vely considered as a rectangle mask. (d) A segmentation mask is generated by GrabCut \cite{rother2004grabcut}. (e) For our method, the supervision is estimated from region proposals (MCG \cite{arbelaez2014multiscale}) by considering bounding box annotations and network feedbacks.}
	\label{fig:masks}
\end{figure*}

\section{Baseline}
\label{sec:baseline}

Our BoxSup method is in general applicable for many existing CNN-based mask-supervised semantic segmentation methods, such as FCN \cite{Long2015}, improvements on FCN \cite{Chen2015,zheng2015conditional}, and others
\cite{Hariharan2015,Dai2015,mostajabi2014feedforward}. In this paper, we adopt our implementation of the FCN method \cite{Long2015} refined by CRF \cite{Chen2015} as the mask-supervised baseline, which we briefly introduce as follows.

The network training of FCN \cite{Long2015} is formulated as a per-pixel regression problem to the ground-truth segmentation masks. Formally, the objective function can be written as:
\begin{equation}
\mathcal{E}(\theta) = \sum_{p} e(X_{\theta}(p), l(p)),
\label{eq:fcn_loss}
\end{equation}
where $p$ is a pixel index, $l(p)$ is the ground-truth semantic label at a pixel, and $X_{\theta}(p)$ is the per-pixel labeling produced by the fully convolutional network with parameters $\theta$.
$e(X_{\theta}(p), l(p))$ is the per-pixel loss function. The network parameters $\theta$ are updated by back-propagation and stochastic gradient descent (SGD). A CRF is used to post-process the FCN results \cite{Chen2015}.

The objective function in Eqn.(\ref{eq:fcn_loss}) demands pixel-level segmentation masks $l(p)$ as supervision. It is not directly applicable if only bounding box annotations are given as supervision. Next we introduce our method for addressing this problem.

\section{Approach}

\subsection{\fontsize{9.9pt}{1em}\selectfont{\textbf{Unsupervised Segmentation for Supervised Training}}}

To harness the bounding boxes annotations, it is desired to estimate segmentation masks from them. This is a widely studied supervised image segmentation problem, and can be addressed by, \eg, GrabCut \cite{rother2004grabcut}. But GrabCut can only generate one or a few samples from one box, which may be insufficient for deep network training.

We propose to generate a set of candidate segments using \emph{unsupervised} region proposal methods (\eg, Selective Search \cite{uijlings2013selective}) due to their nice properties. First, region proposal methods have high recall rates \cite{arbelaez2014multiscale} of having a good candidate in the proposal pool. Second, region proposal methods generate candidates of greater variance, which provide a kind of data augmentation \cite{krizhevsky2012imagenet} for network training. We will show by experiments the improvements of these properties.

The candidate segments are used to update the deep convolutional network. The semantic features learned by the network are then used to pick better candidates. This procedure is iterated.
We formulate this procedure as an objective function as we will describe below.

It is worth noticing that the region proposal is only used for networking training. For inference, the trained FCN is directly applied on the image and produces pixel-wise predictions. So our usage of region proposals does not impact the test-time efficiency.

\begin{figure*}[t]
	\centering
	\includegraphics[width=0.9\linewidth]{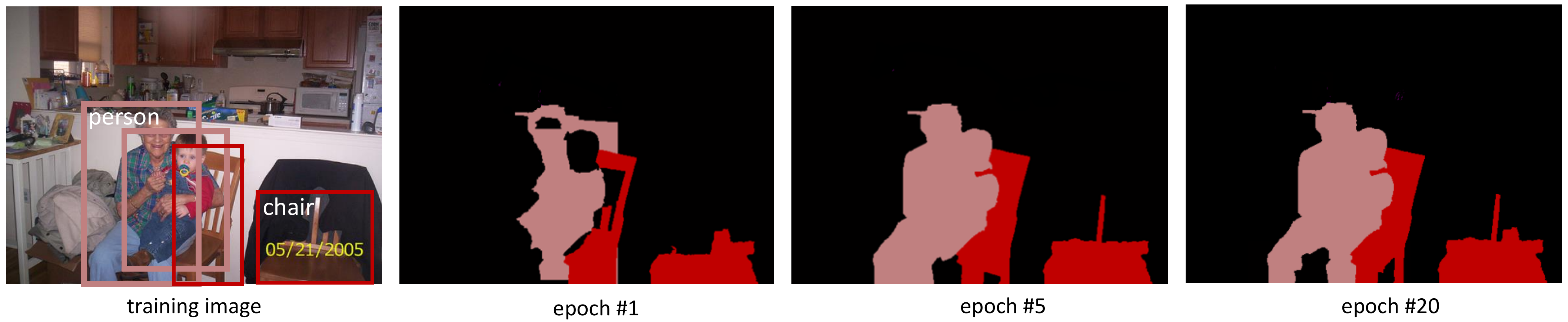}
	\caption{Update of segmentation masks during training. Here we show the masks in epoch \#1, epoch \#5, and epoch \#20. Each segmentation mask will be used as the supervision for the next epoch.}
	\label{fig:epochs}
\end{figure*}

\subsection{Formulation}

As a pre-processing, we use a region proposal method to generate segmentation masks. We adopt Multiscale Combinatorial Grouping (MCG) \cite{arbelaez2014multiscale} by default, while other methods \cite{uijlings2013selective,krahenbuhl2014geodesic} are also evaluated. The proposal candidate masks are fixed throughout the training procedure. But during training, each candidate mask will be assigned a label which can be a semantic category or background. The labels assigned to the masks will be updated.

With a ground-truth bounding box annotation, we expect it to pick out a candidate mask that overlaps the box as much as possible. Formally, we define an overlapping objective function $\mathcal{E}_o$ as:
\begin{equation}\label{eq:ov}
\mathcal{E}_o = \frac{1}{N}\sum_{S}(1-\text{IoU}(B, S))\delta(l_B, l_S).
\end{equation}
Here $S$ represents a candidate segment mask, and $B$ represents a ground-truth bounding box annotation. $\text{IoU}(B, S)\in[0,1]$ is the intersection-over-union ratio computed from the ground-truth box $B$ and the tight bounding box of the segment $S$. The function $\delta$ is equal to one if the semantic label $l_S$ assigned to segment $S$ is the same as the ground-truth label $l_B$ of the bounding box $B$, and zero otherwise. Minimizing $\mathcal{E}_o$ favors higher IoU scores when the semantic labels are consistent. This objective function is normalized by the number of candidate segments $N$.

With the candidate masks and their estimated semantic labels, we can supervise the deep convolutional network as in Eqn.(\ref{eq:fcn_loss}). Formally, we consider the following regression objective function $\mathcal{E}_r$:
\begin{equation}\label{eq:reg}
\mathcal{E}_r = \sum_{p} e(X_{\theta}(p), l_S(p)).
\end{equation}
Here $l_S$ is the estimated semantic label used as supervision for the network training. This objective function is the same as Eqn.(\ref{eq:fcn_loss}) except that its regression target is the estimated candidate segment.

We minimize an objective function that combines the above two terms:
\begin{equation}\label{eq:obj}
\min_{\theta,\{l_S\}} \sum_{i}(\mathcal{E}_o+\lambda\mathcal{E}_r)
\end{equation}
Here the summation $\sum_{i}$ runs over the training images, and $\lambda=3$ is a fixed weighting parameter. The variables to be optimized are the network parameters $\theta$ and the labeling $\{l_S\}$ of all candidate segments $\{S\}$. If only the term $\mathcal{E}_o$ exists, the optimization problem in Eqn.(\ref{eq:obj}) trivially finds a candidate segment that has the largest IoU score with the box; if only the term $\mathcal{E}_r$ exists, the optimization problem in Eqn.(\ref{eq:obj}) is equivalent to FCN. Our formulation simultaneously considers both cases.

\subsection{Training Algorithm}
\label{sec:alg}

The objective function in Eqn.(\ref{eq:obj}) involves a problem of assigning labels to the candidate segments. Next we propose a greedy iterative solution to find a local optimum.

With the network parameters $\theta$ fixed, we update the semantic labeling $\{l_S\}$ for all candidate segments. In our implementation, we only consider the case in which one ground-truth bounding box can ``activate'' (\ie, assign a non-background label to) one and only one candidate. As such, we can simply update the semantic labeling by selecting a single candidate segment for each ground-truth bounding box, such that its cost $\mathcal{E}_o+\lambda\mathcal{E}_r$ is the smallest among all candidates. The selected segment is assigned the ground-truth semantic label associated with that bounding box. All other pixels are assigned the background label.

The above winner-takes-all selection tends to repeatedly use the same or very similar candidate segments, and the optimization procedure may be trapped in poor local optima. To increase the sample variance for better stochastic training, we further adopt a random sampling method to select the candidate segment for each ground-truth bounding box. Instead of selecting the single segment with the largest cost $\mathcal{E}_o+\lambda\mathcal{E}_r$, we randomly sample a segment from the first $k$ segments with the largest costs. In this paper we use $k=5$. This random sampling strategy improves the accuracy by about 2\% on the validation set.

With the semantic labeling $\{l_S\}$ of all candidate segments fixed, we update the network parameters $\theta$. In this case, the problem becomes the FCN problem \cite{Long2015} as in Eqn.(\ref{eq:fcn_loss}). This problem is minimized by SGD.

We iteratively perform the above two steps, fixing one set of variables and solving for the other set.
For each iteration, we update the network parameters using one training epoch (\ie, all training images are visited once), and after that we update the segment labeling of all images. Fig.\ref{fig:epochs} shows the gradually updated segmentation masks during training.
The network is initialized by the model pre-trained in the ImageNet classification dataset, and our algorithm starts from the step of updating segment labels.

Our method is applicable for the semi-supervised case (the ground-truth annotations are mixtures of segmentation masks and bounding boxes). The labeling $l(p)$ is given by candidate proposals as above if a sample only has ground-truth boxes, and is simply assigned as the true label if a sample has ground-truth masks.

In the SGD training of updating the network, we use a mini-batch size of 20, following \cite{Long2015}. The learning rate is initialized to be 0.001 and divided by 10 after every 15 epochs. The training is terminated after 45 epochs.


\section{Experiments}

In all our experiments, we use the publicly released VGG-16 model\footnote{\url{www.robots.ox.ac.uk/~vgg/research/very_deep/}} \cite{Simonyan2015} that is pre-trained on ImageNet \cite{Russakovsky2014}. The VGG model is also used by all competitors \cite{Long2015,Hariharan2015,Dai2015,Chen2015,mostajabi2014feedforward} compared in this paper.

\subsection{Experiments on PASCAL VOC 2012}

We first evaluate our method on the PASCAL VOC 2012 semantic segmentation benchmark \cite{everingham2010pascal}. This dataset involves 20 semantic categories of objects. We use the ``comp6'' evaluation protocol.
The accuracy is evaluated by mean IoU scores.
The original training data has 1,464 images.
Following \cite{hariharan2011semantic}, the training data with ground-truth segmentation masks are augmented to 10,582 images. The validation and test sets have 1,449 and 1,456 images respectively. When evaluating the validation set or the test set, we only use the training set for training.
A held-out 100 random validation images are used for cross-validation to set hyper-parameters.

\vspace{8pt}
\noindent\textbf{Comparisons of Supervision Strategies}

\setlength{\tabcolsep}{5pt}
\begin{table}[t]
	\renewcommand{\arraystretch}{1.1}
	\begin{center}
		\small
		\begin{tabular}{x||x|x|x|x|x}
			\hline
			data & \multicolumn{3}{ c| }{VOC train} & \multicolumn{2}{ c }{VOC train + COCO}\\
			\hline
			total \# & \multicolumn{3}{ c| }{10,582} & \multicolumn{2}{ c }{133,869}\\\hline
			\tn{supervision} & mask & box  & semi & mask & semi\\
			\hline
			mask \# & 10,582 & - & 1,464 & 133,869 & 10,582\\
			box \# & - & 10,582 & 9,118 & - & 123,287\\
			\hline
			mean IoU & 63.8 & 62.0 & 63.5 & 68.1 & 68.2\\
			\hline
		\end{tabular}
	\end{center}
	\caption{Comparisons of supervision in PASCAL VOC 2012 validation.}
	\label{tab:voc2012_val_supervision}
\end{table}

Table~\ref{tab:voc2012_val_supervision} compares the results of using different strategies of supervision on the validation set. When all ground-truth masks are used as supervision, the result is our implementation of the baseline DeepLab-CRF \cite{Chen2015}.
Our reproduction has a score of 63.8 (Table~\ref{tab:voc2012_val_supervision}, ``mask only''), which is very close to 63.74 reported in \cite{Chen2015} under the same setting. So we believe that our reproduced baseline is convincing.

When all 10,582 training samples are replaced with bounding box annotations, our method yields a score of 62.0 (Table~\ref{tab:voc2012_val_supervision}, ``box only''). Though the supervision information is substantially weakened, our method shows a graceful degradation (1.8\%) compared with the strongly supervised baseline of 63.8. This indicates that in practice we can avoid the expensive mask labeling effort by using only bounding boxes, with small accuracy loss.

Table~\ref{tab:voc2012_val_supervision} also shows the semi-supervised result of our method. This result uses the ground-truth masks of the original 1,464 training images and the bounding box annotations of the rest 9k images. The score is 63.5 (Table~\ref{tab:voc2012_val_supervision}, ``semi''), on par with the strongly supervised baseline. Such semi-supervision replaces 9/10 of the segmentation mask annotations with bounding box annotations. This means that we can greatly reduce the labeling effort by dominantly using bounding box annotations.

\begin{figure*}[t]
	\centering
	\includegraphics[width=0.85\linewidth]{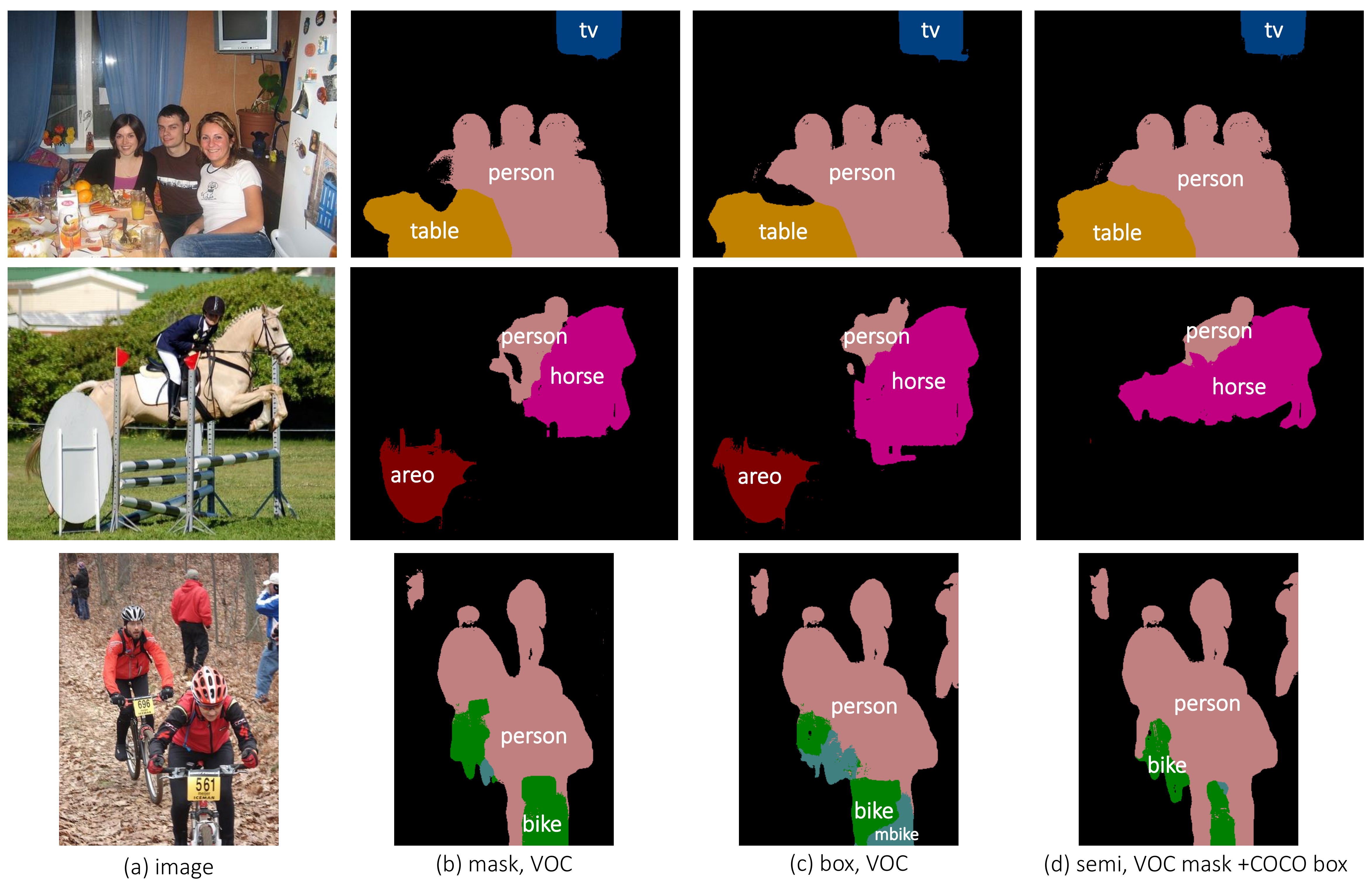}
	\caption{Example semantic segmentation results on \textbf{PASCAL VOC 2012} validation using our method. (a) Images. (b) Supervised by masks in VOC. (c) Supervised by boxes in VOC. (d) Supervised by masks in VOC and boxes in COCO.}
	\label{fig:results}
\end{figure*}

As a proof of concept, we further evaluate using a substantially larger set of boxes. We use the Microsoft COCO dataset \cite{Lin2014} that has 123,287 images with available ground-truth segmentation masks. This dataset has 80 semantic categories, and we only use the 20 categories that also present in PASCAL VOC. For our mask-supervised baseline, the result is a score of 68.1 (Table~\ref{tab:voc2012_val_supervision}). Then we replace the ground-truth segmentation masks in COCO with their tight bounding boxes. Our semi-supervised result is 68.2 (Table~\ref{tab:voc2012_val_supervision}), on par with the strongly supervised baseline. Fig.~\ref{fig:results} shows some visual results in the validation set.

The semi-supervised result (68.2) that uses VOC+COCO is considerably better than the strongly supervised result (63.8) that uses VOC only. The 4.4\% gain is contributed by the extra large-scale bounding boxes in the 123k COCO images. This comparison suggests a promising strategy - we may make use of the larger amount of existing bounding boxes annotations to improve the overall semantic segmentation results, as further analyzed below.

\vspace{8pt}
\noindent\textbf{Error Analysis}

\begin{figure}[t]
\centering
\includegraphics[width=0.9\linewidth]{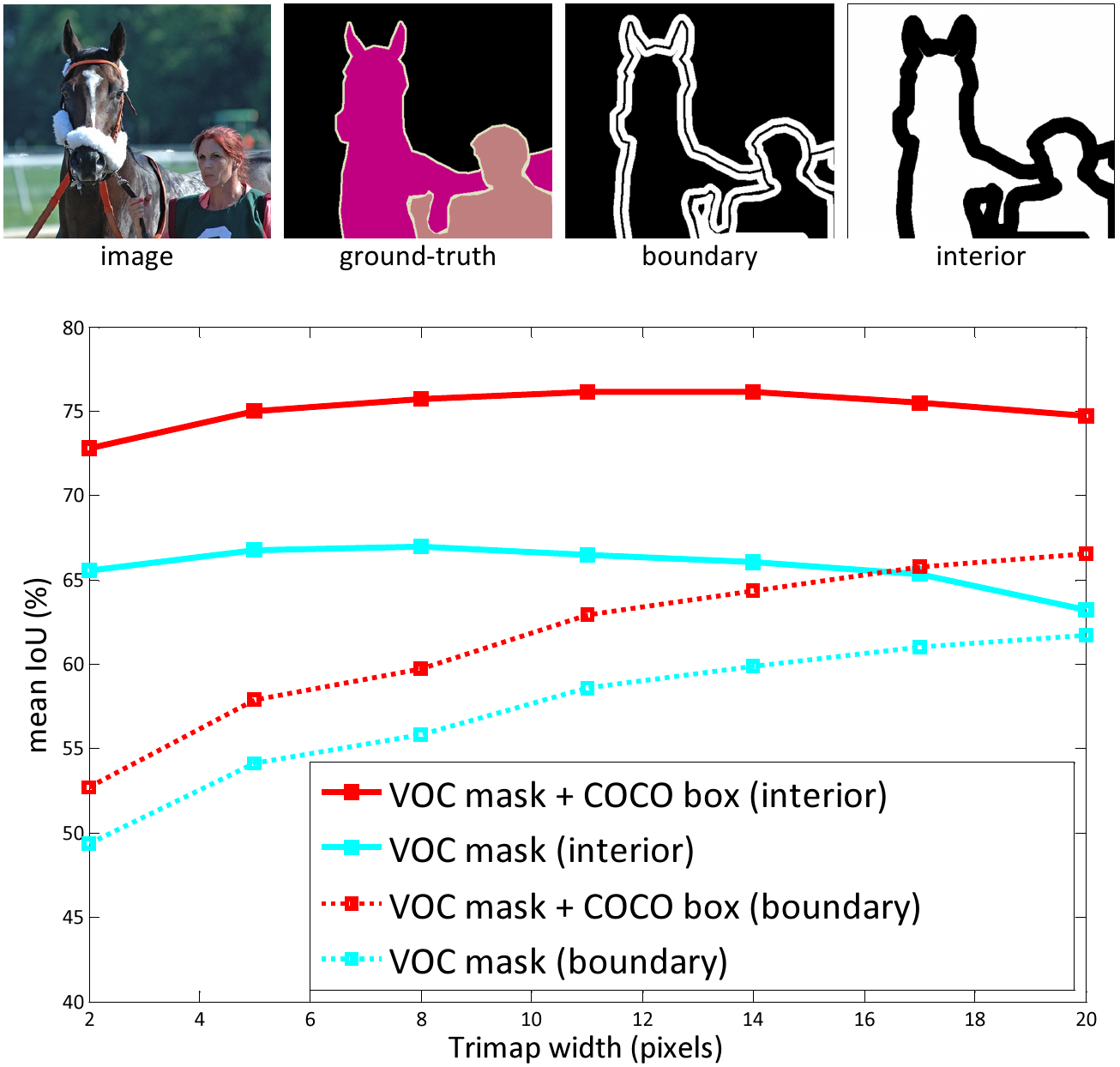}
\caption{Error analysis on the validation set. Top: (from left to right) image, ground-truth, \emph{boundary} regions marked as white, \emph{interior} regions marked as white). Bottom: \emph{boundary} and \emph{interior} mean IoU, using VOC masks only (blue) and using extra COCO boxes (red).}
\label{fig:trimap}
\end{figure}

Why can a large set of bounding boxes help improve convolutional networks? The error in semantic segmentation can be roughly thought of as two types: (i) \emph{recognition} error that is due to confusions of recognizing object categories, and (ii) \emph{boundary} error that is due to misalignments of pixel-level labels on object boundaries. Although the bounding box annotations have no information about the object boundaries, they provide extra object instances for recognizing them. We may expect that the large amount of boxes mainly improve the recognition accuracy.

To analyze the error, we separately evaluate the performance on the \emph{boundary} regions and \emph{interior} regions. Following \cite{Kohli2009,Chen2015}, we generate a ``trimap'' near the ground-truth boundaries (Fig.~\ref{fig:trimap}, top). We evaluate mean IoU scores inside/outside the bands, referred to as boundary/interior regions. Fig.~\ref{fig:trimap} (bottom) shows the results of using different band widths for the trimaps.

For the interior region, the accuracy of using the extra COCO boxes (red solid line, Fig.~\ref{fig:trimap}) is considerably higher than that of using VOC masks only (blue solid line). On the contrary, the improvement on the boundary regions is relatively smaller (red dash line \vs blue dash line). Note that correctly recognizing the interior may also help improve the boundaries (\eg, due to the CRF post-processing). So the improvement of the extra boxes on the boundary regions is secondary.

Because the accuracy in the interior region is mainly determined by correctly recognizing objects, this analysis suggests that the large amount of boxes improve the feature quality of a learned BoxSup model for better recognition.

\setlength{\tabcolsep}{8pt}
\renewcommand{\arraystretch}{1.05}
\begin{table}[t]
\small
	\begin{center}
		\begin{tabular}{x|x}
\hline
masks & mean IoU\\
\hline
\hline
rectangles & 52.3\\
GrabCut & 55.2\\
WSSL \cite{papandreou2015weakly} & 58.5\\
\hline
ours w/o sampling & 59.7\\
ours & \underline{62.0}\\
\hline
		\end{tabular}
	\end{center}
	\caption{Comparisons of estimated masks for supervision in PASCAL VOC 2012 validation. All methods only use 10,582 bounding boxes as annotations, with no ground-truth segmentation mask used.}
	\label{tab:voc2012_val_approaches}
\end{table}

\setlength{\tabcolsep}{8pt}
\renewcommand{\arraystretch}{1.1}
\begin{table}[t]
	\begin{center}
		\begin{tabular}{x|x|x|x}
			\hline
 & SS & GOP & MCG \\
\hline
mean IoU & 59.5 & 60.4 & \underline{62.0}\\
\hline
		\end{tabular}
	\end{center}
	\caption{Comparisons of the effects of region proposal methods on our method in PASCAL VOC 2012 validation. All methods only use 10,582 bounding boxes as annotations, with no ground-truth segmentation mask used.}
	\label{tab:voc2012_val_proposals}
\end{table}

\vspace{8pt}
\noindent\textbf{Comparisons of Estimated Masks for Supervision}

In Table~\ref{tab:voc2012_val_approaches} we evaluate different methods of estimating masks from bounding boxes for supervision. As a na\"{\i}ve baseline, we fill each bounding box with its semantic label, and consider it as a rectangular mask (Fig.~\ref{fig:masks}(c)). Using these rectangular masks as the supervision throughout training, the score is 52.3 on the validation set. We also use GrabCut \cite{rother2004grabcut} to generate segmentation masks from boxes (Fig.~\ref{fig:masks}(d)). With the GrabCut masks as the supervision throughout training, the score is 55.2. In both cases, the masks are not updated by the network feedbacks.

Our method has a score 62.0 (Table~\ref{tab:voc2012_val_approaches}) using the same set of bounding box annotations. This is a considerable gain over the baseline using fixed GrabCut masks. This indicates the importance of the mask quality for supervision. Fig.~\ref{fig:epochs} shows that our method iteratively updates the masks by the network, which in turn improves the network training.

We also evaluate a variant of our method where each time the updated mask is the candidate with the largest cost, instead of randomly sampled from the first $k$ candidates (see Sec.~\ref{sec:alg}). This variant has a lower score of 59.7 (Table~\ref{tab:voc2012_val_approaches}). The random sampling strategy, which is data augmentation and increases sample variances, is beneficial for training.

Table~\ref{tab:voc2012_val_approaches} also shows the result of the concurrent method WSSL \cite{Chen2015} under the same evaluation setting. Its results is 58.5. This result suggests that our method estimates more accurate masks than \cite{Chen2015} for supervision.

\vspace{8pt}
\noindent\textbf{Comparisons of Region Proposals}

Our method resorts to unsupervised region proposals for training.
In Table \ref{tab:voc2012_val_proposals}, we compare the effects of various region proposals on our method: Selective Search (SS) \cite{uijlings2013selective}, Geodesic Object Proposals (GOP) \cite{krahenbuhl2014geodesic}, and MCG \cite{arbelaez2014multiscale}.
Table \ref{tab:voc2012_val_proposals} shows that MCG \cite{arbelaez2014multiscale} has the best accuracy, which is consistent with its segmentation quality evaluated by other metrics in \cite{arbelaez2014multiscale}. Note that at test-time our method does not need region proposals. So the better accuracy of using MCG implies that
our method effectively makes use of the higher quality segmentation masks to train a better network.

\vspace{8pt}
\noindent\textbf{Comparisons on the Test Set}

\setlength{\tabcolsep}{4pt}
\renewcommand{\arraystretch}{1.1}
\begin{table}[t]
	\begin{center}
\begin{small}
		\begin{tabular}{x|x|r|r|x}
			\hline
			method & sup. & mask \# & box \# & mIoU\\
			\hline
			\hline
			FCN  \cite{Long2015} & mask & V 10k & - & 62.2\\
			\tn{DeepLabCRF}  \cite{Chen2015} & mask & V 10k & - & 66.4\\
			WSSL \cite{papandreou2015weakly} & box & - & V 10k & 60.4\\
			\textbf{BoxSup} & box & - & V 10k  & 64.6\\
			\textbf{BoxSup} & semi & V 1.4k & V 9k & 66.2\\
			\hline
			WSSL \cite{papandreou2015weakly} & mask & V+C 133k & - & 70.4\\
			\textbf{BoxSup} & semi & V 10k  & C 123k & 71.0\\
			\textbf{BoxSup} & semi & V 10k  & V$_{07}$+C 133k & 73.1 \\
			\textbf{BoxSup+} & semi & V 10k  & V$_{07}$+C 133k & \textbf{75.2}\\
            \hline
		\end{tabular}
\end{small}
	\end{center}
	\caption{Results on \textbf{PASCAL VOC 2012 test} set. In the supervision (``sup'') column, ``mask'' means all training samples are with segmentation mask annotations, ``box'' means all training samples are with bounding box annotations, and ``semi'' means mixtures. ``V'' denotes the VOC data, ``C'' denotes the COCO data, and ``V$_{07}$'' denotes the VOC 2007 data which only has bounding boxes available.}
	\label{tab:voc2012_test}
\end{table}

Next we compare with the state-of-the-art methods on the PASCAL VOC 2012 {\em test} set. In Table~\ref{tab:voc2012_test}, the methods are based on the same FCN baseline and thus fair comparisons are made to evaluate the impact of mask/box/semi-supervision.

As shown in Table \ref{tab:voc2012_test}, our \emph{box-supervised} result that only uses VOC bounding boxes is 64.6. This compares favorably with the WSSL \cite{papandreou2015weakly} counterpart (60.4) under the same setting. On the other hand, our \emph{box-supervised} result has a graceful degradation (1.8\%) compared with the \emph{mask-supervised} DeepLab-CRF (66.4 \cite{Chen2015}) using the VOC training data.
Moreover, our semi-supervised variant which replaces 9/10 segmentation mask annotations with bounding boxes has a score of 66.2. This is on par with the mask-supervised counterpart of DeepLab-CRF, but the supervision information used by our method is much weaker.

In the WSSL paper \cite{papandreou2015weakly}, by using all segmentation mask annotations in VOC and COCO, the strongly mask-supervised result is 70.4. Our semi-supervised method shows a higher score of \textbf{71.0}. Remarkably, our result uses the bounding box annotations from the 123k COCO images. So our method has a more accurate result but uses much weaker annotations than \cite{papandreou2015weakly}.

On the other hand, compared with the DeepLab-CRF result (66.4), our method has a 4.6\% gain enjoyed from exploiting the \emph{bounding box} annotations of the COCO dataset. This comparison demonstrates the power of our method that exploits large-scale bounding box annotations to improve accuracy.

\vspace{8pt}
\noindent\textbf{Exploiting Boxes in PASCAL VOC 2007}

To further demonstrate the effect of BoxSup, we exploit the bounding boxes in the PASCAL VOC 2007 dataset \cite{everingham2010pascal}. This dataset has no mask annotations. It is a de facto dataset which mask-supervised methods are \emph{not} able to use.

We exploit all 10k images in the VOC 2007 trainval and test sets. We train a BoxSup model using the union set of VOC~2007 boxes, COCO boxes, and the augmented VOC 2012 training set. The score improves from 71.0 to \textbf{73.1} (Table~\ref{tab:voc2012_test}) because of the extra box training data. It is reasonable for us to expect further improvement if more bounding box annotations are available.

\begin{figure*}[t]
	\centering
	\includegraphics[width=0.9\linewidth]{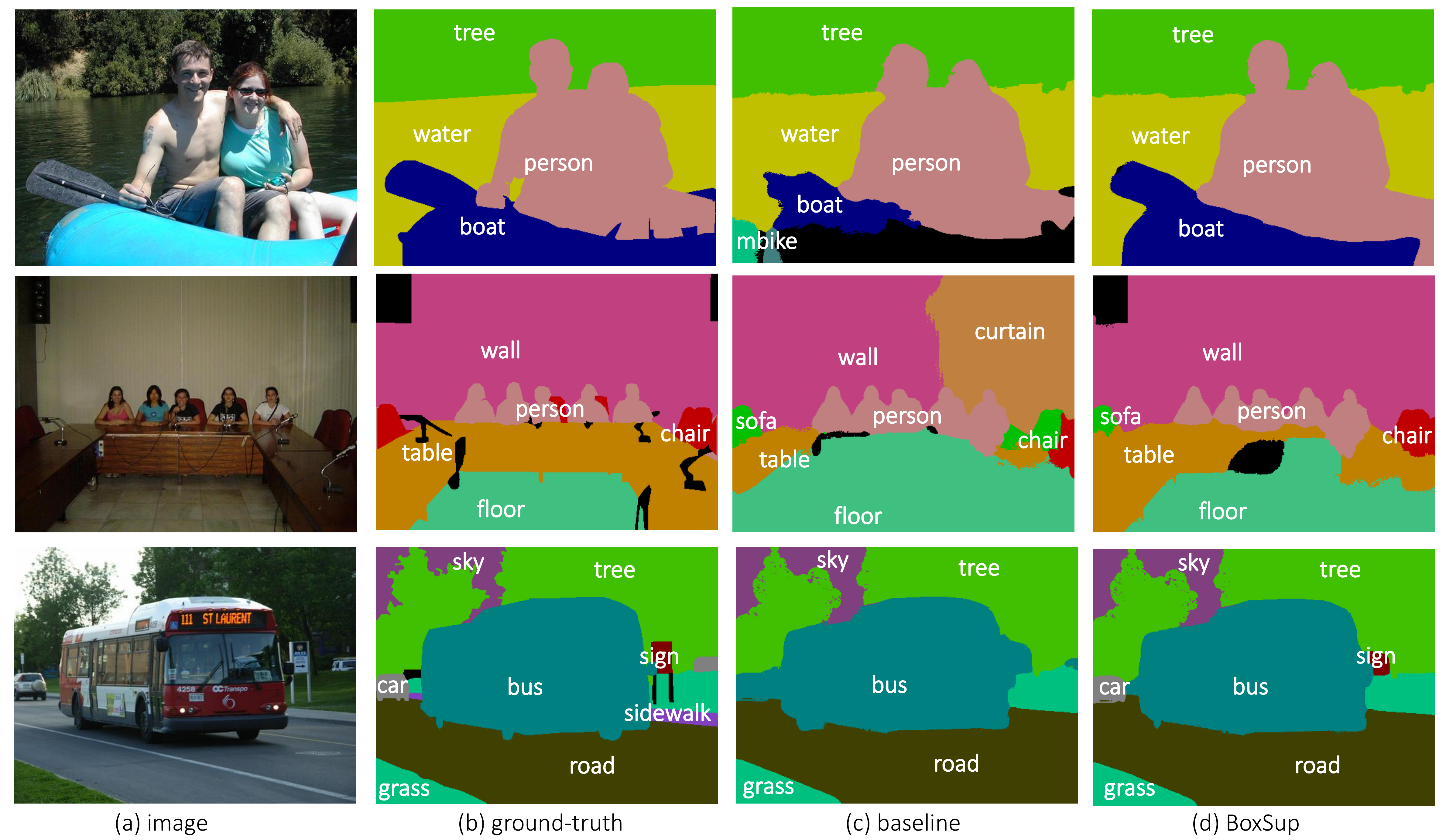}
	\caption{Example results on \textbf{PASCAL-CONTEXT} validation. (a) Images. (b) Results of our baseline (35.7 mean IoU), trained using VOC masks. (c) Results of BoxSup (40.5 mean IoU), trained using VOC masks and COCO boxes.}
	\label{fig:results_voc2010}
\end{figure*}

\vspace{8pt}
\noindent\textbf{Baseline Improvement}

Although our focus is mainly on exploiting boxes as supervision, it is worth noticing that our method may also benefit from other improvements on the mask-sup baseline (FCN in our case). Concurrent with our work, there are a series of improvements \cite{zheng2015conditional,Chen2015} made on FCN, which achieve excellent results using strong mask-supervision from VOC and COCO data.

To show the potential of our BoxSup method in parallel with improvements on the baseline, we use a simple test-time augmentation to boost our results. Instead of computing pixel-wise predictions on a single scale, we compute the score maps from two extra scales ($\pm20\%$ of the original image size) and bilinearly re-scale the score maps to the original size. The scores from three scales are averaged. This simple modification boosts our result from 73.1 to \textbf{75.2} (BoxSup+, Table~\ref{tab:voc2012_test}) in the VOC 2012 test set. This result is on par with the latest results using strong mask-supervision from both VOC and COCO, but in our case the  COCO dataset only provides bounding boxes.

\setlength{\tabcolsep}{6pt}
\renewcommand{\arraystretch}{1.1}
\begin{table}[t]
	\begin{center}
		\begin{tabular}{x|x|x|x|x}
			\hline
			method & sup. & mask \# & box \# & \tn{mean IoU}\\
			\hline
			\hline
			O$_2$P \cite{carreira2012semantic} & mask & V 5k & - & 18.1\\
			CFM \cite{Dai2015} & mask & V 5k & - & 34.4\\
			FCN \cite{Long2015} & mask & V 5k & - & 35.1\\
			\hline
			baseline & mask & V 5k & - & 35.7\\
			\textbf{BoxSup} & semi & V 5k & C 123k & \textbf{40.5}\\
			\hline
		\end{tabular}
	\end{center}
	\caption{Results on \textbf{PASCAL-CONTEXT} \cite{mottaghi2014role} validation. Our baseline is our implementation of FCN+CRF. ``V'' denotes the VOC data, and ``C'' denotes the COCO data.}
	\label{tab:voc2010_val}
\end{table}

\subsection{Experiments on PASCAL-CONTEXT}

We further perform experiments on the recently labeled PASCAL-CONTEXT dataset \cite{mottaghi2014role}. This dataset provides ground-truth semantic labels for the whole scene, including object and stuff (\eg, grass, sky, water). Following the protocol in \cite{mottaghi2014role,Dai2015,Long2015}, the semantic segmentation is performed on the most frequent 59 categories (identified by \cite{mottaghi2014role}) plus a background category. The accuracy is measured by mean IoU scores. The training and evaluation are performed on the training and validation sets that have 4,998 and 5,105 images respectively.

To train a BoxSup model for this dataset, we first use the \emph{box annotations} from all 80 object categories in the COCO dataset to train the FCN (using VGG-16). This network ends with an 81-way (with an extra one for background) layer. Then we remove this last layer and add a new 60-way layer for the 59 categories of PASCAL-CONTEXT. We fine-tune this model in the 5k training images of PASCAL-CONTEXT. A CRF for post-processing is also used. We do no use the test-time scale augmentation.

Table \ref{tab:voc2010_val} shows the results in PASCAL-CONTEXT. The methods of CFM \cite{Dai2015} and FCN \cite{Long2015} are both based on the VGG-16 model. Our baseline method, which is our implementation of FCN+CRF,
has a score of 35.7 using masks of the 5k training images. Using our BoxSup model pre-trained using the COCO boxes, the result is improved to \textbf{40.5}. The 4.8\% gain is solely because of the bounding box annotations in COCO that improve our network training. Fig.~\ref{fig:results_voc2010} shows some examples of our results for joint object and stuff segmentation.

\section{Conclusion}

The proposed BoxSup method can effectively harness bounding box annotations to train deep networks for semantic segmentation. Our BoxSup method that uses 133k bounding boxes and 10k masks achieves state-of-the-art results.
Our error analysis suggests that semantic segmentation accuracy is hampered by the failure of recognizing objects, which large-scale data may help with.

{\small
\bibliographystyle{ieee}
\bibliography{boxsup_v3}
}

\end{document}